\documentclass[letterpaper]{article} 
\usepackage{aaai25}  
\usepackage{times}  
\usepackage{helvet}  
\usepackage{courier}  
\usepackage[hyphens]{url}  
\usepackage{graphicx} 
\usepackage{natbib}  
\usepackage{caption} 
\usepackage{algorithm}
\usepackage{algorithmic}
\usepackage{url}            
\usepackage{booktabs}       
\usepackage{amsfonts}       
\usepackage{amsmath}
\usepackage{nicefrac}       
\usepackage{microtype}      
\usepackage{xcolor}         
\usepackage{graphbox}
\usepackage{bm}
\usepackage{siunitx}
\usepackage{caption}
\newcommand{\TODO}[1]{}
\renewcommand{\TODO}[1]{{\color{cyan} [TODO: {#1}]}}
\newcommand{\WIP}[1]{}
\renewcommand{\WIP}[1]{{\color{magenta} [WIP: {#1}]}}

\usepackage{newfloat}
\usepackage{listings}
\DeclareCaptionStyle{ruled}{labelfont=normalfont,labelsep=colon,strut=off} 
\floatstyle{ruled}
\newfloat{listing}{tb}{lst}{}
\floatname{listing}{Listing}
\title{Explanation Bottleneck Models}
\author{
    Shin'ya Yamaguchi\textsuperscript{\rm 1,2}\thanks{shinya.yamaguchi@ntt.com} and Kosuke Nishida\textsuperscript{\rm 1}
}
\affiliations{
    \textsuperscript{\rm 1}NTT, 
    \textsuperscript{\rm 2}Kyoto University\\
}

\usepackage{bibentry}

\begin{document}

\maketitle

\begin{abstract}
Recent concept-based interpretable models have succeeded in providing meaningful explanations by pre-defined concept sets. However, the dependency on the pre-defined concepts restricts the application because of the limited number of concepts for explanations. This paper proposes a novel interpretable deep neural network called \textit{explanation bottleneck models} (XBMs). XBMs generate a text explanation from the input without pre-defined concepts and then predict a final task prediction based on the generated explanation by leveraging pre-trained vision-language encoder-decoder models. To achieve both the target task performance and the explanation quality, we train XBMs through the target task loss with the regularization penalizing the explanation decoder via the distillation from the frozen pre-trained decoder. Our experiments, including a comparison to state-of-the-art concept bottleneck models, confirm that XBMs provide accurate and fluent natural language explanations without pre-defined concept sets. Code will be available at \url{https://github.com/yshinya6/xbm/}.
\end{abstract}

%

\section{Introduction}
Although deep learning models can achieve remarkable performance on many applications, they are black-box, i.e., their output predictions are not interpretable for humans.
Introducing concept bottleneck models (CBMs, \citet{Koh_ICML20_concept_bottleneck}) is a promising approach to interpreting the output of deep models.
In contrast to black-box models that directly predict output labels from input in an end-to-end fashion, CBMs first predict \textit{concept} labels from input and then predict final target class labels from the predicted concepts.
Since the predicted concepts represent semantic input ingredients, this two-staged prediction enables users to know the reasons for the final target label predictions and interactively intervene in the decision-making process for critical applications such as healthcare~\cite{Chauhan_AAAI23_interactive_CBM}.

However, the existing CBMs depend on the fixed pre-defined concept sets to predict final labels.
In other words, they can not provide interpretability to any other than the pre-defined concepts.
We argue that this limitation presents a fundamental challenge for CBMs in achieving interpretable deep models.
Although recent CBM variants leveraging pre-trained large language models~\cite{Yuksekgonul_ICLR23_post-hoc_CBMs,Oikarinen_ICLR23_label-free_CBMs} enable to express concepts of arbitrary target classes, the interpretability is still restricted to a fixed and small number of concepts.
This is because a large number of concept labels are difficult to learn due to their long-tail distribution and are less interpretable by the limitation of human perception~\cite{Ramaswamy_CVPR23_overlooked_factors_CBMs}.
In fact, the prior works restrict the number of concepts by filtering with the similarity between concepts and training images to maintain the performance and interpretability~\cite{Oikarinen_ICLR23_label-free_CBMs,Yang_CVPR23_language_in_a_bottle_CBMs}.
Therefore, as long as they depend on pre-defined concepts, CBMs are restricted in the number of interpretable concepts and are insufficient to explain the output of deep models.

\begin{figure}[t]
    \centering
    \includegraphics[width=\linewidth]{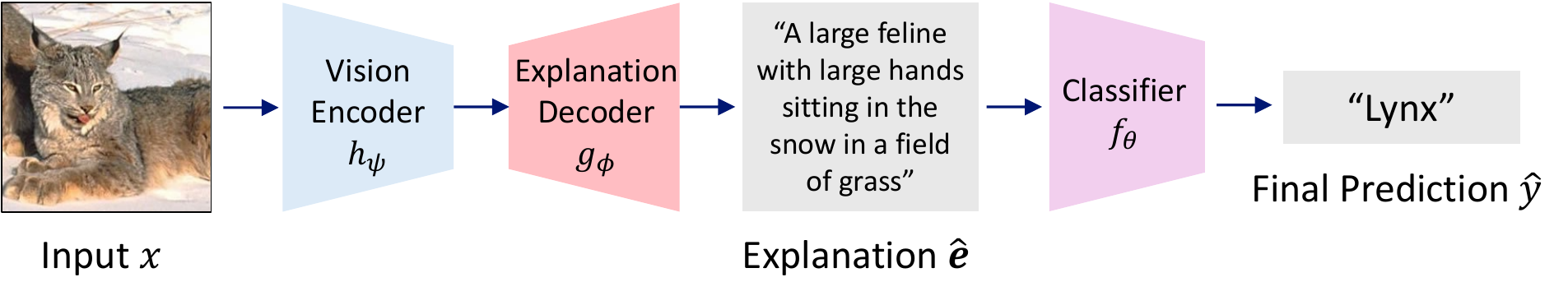}
    \vspace{-2.5mm}
    \caption{
    Explanation bottleneck models (XBMs). We propose an interpretable model that generates text explanations for the input embedding with respect to target tasks and then predicts final task labels from the explanations.
    }
    \label{fig:top}
    \vspace{-2.5mm}
\end{figure}

This paper tackles a research problem where we do not assume pre-defined concept sets for constructing interpretable deep neural networks.
To this end, we propose a novel family of interpretable models called \textit{explanation bottleneck models} (XBMs), which leverage pre-trained multi-modal encoder-decoder models that can generate text descriptions from input data (e.g., BLIP~\cite{Li_ICML22_BLIP,Li_ICML23_BLIP2}).
Leveraging pre-trained multi-modal encoder-decoder enables capturing concepts that actually appeared in the input beyond pre-defined concept sets.
Our key idea is to decode concepts as text explanations from input and then predict the final label with a classifier that takes the decoded explanations (Fig.~\ref{fig:top}).
In contrast to CBMs, which make predictions based on pre-defined concepts, XBMs make predictions based on concepts actually appeared in the input data through the decoded explanations and can provide an intuitive interpretation of the final prediction tied to the input.
Through end-to-end training, XBMs aim to generate explanations focusing on the textual features for solving the target task.

A major challenge for XBMs is forgetting the text generation capability during training on target tasks.
Since target datasets usually lack ground-truth text labels, it is challenging to avoid catastrophic forgetting.
To generate high-quality explanations, we introduce a training technique called \textit{explanation distillation}, which penalizes the text decoders by the reference explanations generated by frozen pre-trained text decoders.
Solving target tasks with explanation distillation enables XBMs to decode explanations from input data in natural sentences without corruption.

We conduct experiments to evaluate XBMs on multiple datasets by comparing them to existing CBMs and black-box baselines regarding interpretability and target task performance.
Our experiments show that XBMs can provide a more relevant explanation to input than the pre-defined concepts of existing CBMs while achieving competitive performance to black-box baselines and largely outperforming CBMs in target test accuracy.
We also show that training XBMs can enhance the multi-modal understanding capability of backbone vision-language models by focusing on the target-related vocabulary.
Further, we confirm the reliability and practicality of the XBMs' explanations through the experiments intervening with the random texts and the ground-truth explanations.

\section{Explanation Bottleneck Models}\label{sec:method}
\begin{figure}[t]
    \centering
    \includegraphics[width=\linewidth]{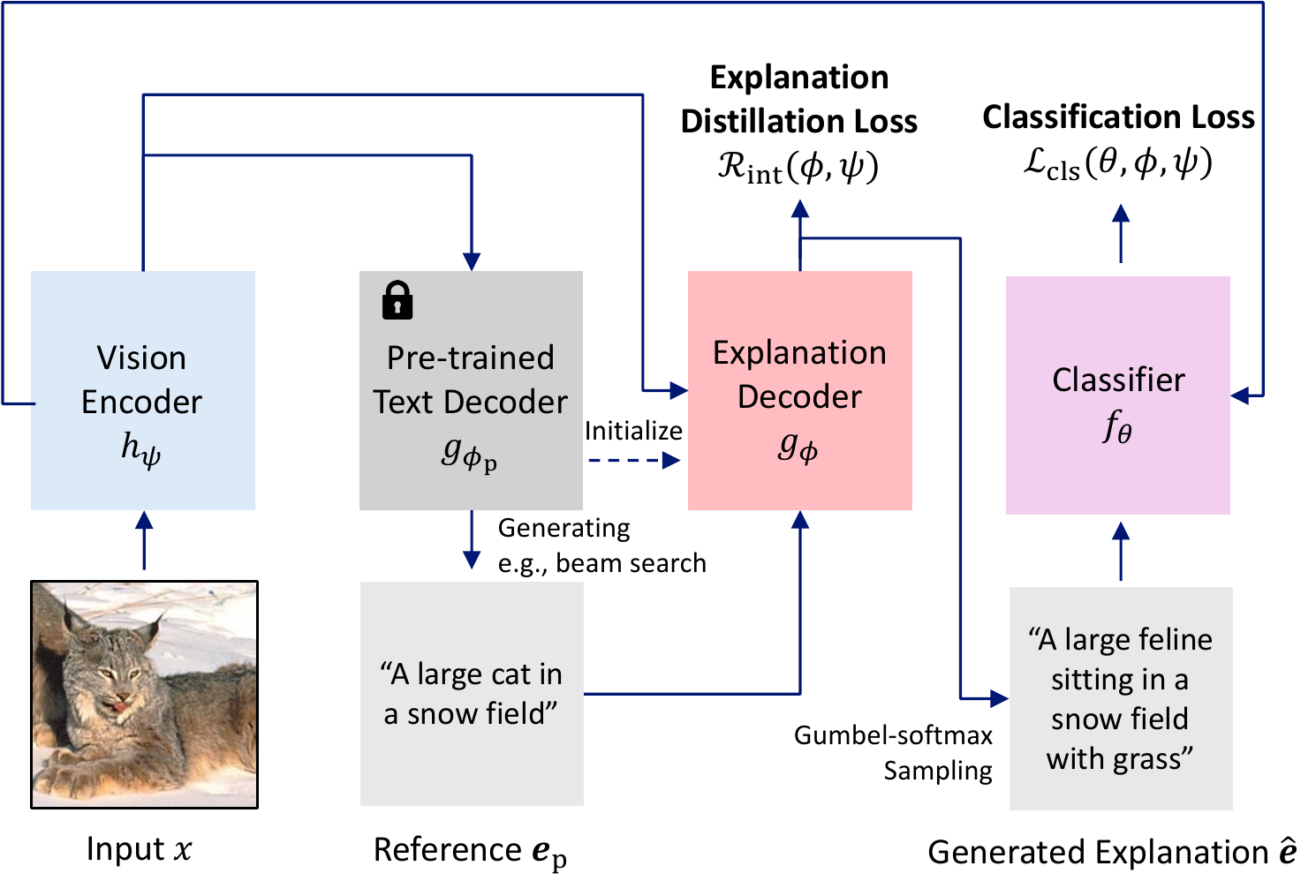}
    \vspace{-2.5mm}
    \caption{
    Training of XBMs. An XBM is optimized by the target task loss with explanation distillation. Explanation distillation leverages a reference explanation \(\bm{e}_\mathrm{p}\) generated from a pre-trained text decoder \(g_{\phi_\mathrm{p}}\) for penalizing the output distribution of an explanation decoder \(g_\phi\) to maintain the interpretable text generation capability of \(g_\phi\).
    }
    \label{fig:method}
    \vspace{-2.5mm}
\end{figure}

This section introduces the principle of explanation bottleneck models (XBMs).
XBMs are interpretable deep learning models that predict a final label from the generated explanation text from XBMs themselves.
Since the predicted final labels are based on the generated explanation of input images, we can naturally interpret the explanation as the reason for the prediction of XBMs.
Figure~\ref{fig:method} illustrates the overview of training an XBM.
An XBM consists of a visual encoder \(h_\psi\), an explanation decoder \(g_\phi\), and a classifier \(f_\theta\) for predicting final target labels.
Among them, \(h_\psi\) and \(g_\phi\) are initialized by an arbitrary pre-trained multi-modal encoder-decoder like BLIP~\cite{Li_ICML22_BLIP}.
\(f_\theta\) is a multi-modal classifier built on a transformer that takes the generated explanations as input and conditions the cross-attention layers with image embeddings; this design is inspired by hybrid post-hoc CBMs~\cite{Yuksekgonul_ICLR23_post-hoc_CBMs} that uses input embeddings to complement missing concepts not in the predicted concepts.
We also confirm the practicality when using a text classifier in Section~\ref{sec:exp_text_cls}.
In this section, we mainly describe XBMs with a multi-modal classifier.
XBMs are trained by the target classification loss in an end-to-end manner.
Since na\"ive training leads to collapse in generated text explanation, we avoid the collapse by \textit{explanation distillation}.
Explanation distillation penalizes the explanation decoder with a reference text generated from a frozen pre-trained text decoder \(g_{\phi_\mathrm{p}}\) to prevent the decoders from forgetting the text generation capability.\looseness-1

\subsection{Problem Setting}
We consider a \(K\)-class image classification task as the target task.
We train neural network models \(h_\psi: \mathcal{X} \to \mathbb{R}^{d_\mathcal{X}}\), \(g_{\phi} :\mathbb{R}^{d_\mathcal{X}} \to \mathcal{E}\), and \(f_{\theta} :(\mathbb{R}^{d_\mathcal{X}}, \mathcal{E}) \to \mathcal{Y}\) on a labeled target dataset \(\mathcal{D}=\{(x^i,y^i) \in \mathcal{X}\times\mathcal{Y}\}^{N}_{i=1}\), where \(\mathcal{X}\), \(\mathcal{E}\), and \(\mathcal{Y}\) are the input, text explanation, and output label spaces, respectively.
The text explanation space consists of token sequences of the length \(L\) with token vocabulary \(\mathcal{V}\), i.e., \(\mathcal{E}=\mathcal{V}^L\).
\(h_\psi\) is a vision encoder, which embeds an input \(x\) into \({d_\mathcal{X}}\) dimensional space, \(g_\phi\) is an auto-regressive text decoder that generates a text explanation \(\bm{e}\in\mathcal{E}\) from an input embedding \(h_\psi(x)\), and \(f_\theta\) is a classifier that predicts a final target task label \(y\).
We assume that \(h_\psi\) and \(g_\phi\) are initialized by pre-trained multi-modal model's parameters \(\psi_\mathrm{p}\) and \(\phi_\mathrm{p}\), which are pre-trained on large-scale text-image paired datasets with an existing method such as BLIP~\cite{Li_ICML22_BLIP} and LLaVA~\cite{liu_NeurIPS23_llava}.
Note that we do not assume ground truth text explanation set \(\{\bm{e}^i\}^N_{i=1}\) in \(\mathcal{D}\) for training \(g_\phi\).\looseness-1

This setting is similar to that of concept bottleneck models (CBMs, \citet{Koh_ICML20_concept_bottleneck}), where a model predicts a final label \(y\) from a set of concepts \(\{c^j\in\mathcal{C}\}^M_{j=1}\) decoded from input \(x\) instead of using \(\bm{e}\).
The major difference is in the assumption of pre-defined concept sets: our setting does not explicitly specify the words and phrases for the explanations, whereas CBMs explain the model's output based on the words and phrases in a pre-defined concept set \(\{c^j\}\).\looseness-1

\subsection{Objective Function}
XBMs aim to achieve high target classification accuracy while providing interpretable explanations of the predictions.
To this end, XBMs solve an optimization problem with a regularization term defined by the following objective function.
\begin{eqnarray}
     &\min\limits_{\theta,\phi,\psi}   \mathcal{L}_\mathrm{cls}(\theta,\phi,\psi)+\lambda\mathcal{R}_\mathrm{int}(\phi, \psi),\label{eq:obj_xbm}\\
    &\mathcal{L}_\mathrm{cls}(\theta,\phi,\psi) = {\mathbb{E}_{(x,y)\in \mathcal{D}}}~\ell_\mathrm{CE}(f_\theta\circ g_\phi \circ h_\psi(x),y),\label{eq:cls_loss}
\end{eqnarray}
where \(\mathcal{R}_\mathrm{int}(\cdot)\) is a regularization term that guarantees the fluency of the explanations generated from \(g_\phi\), \(\lambda\) is a hyperparameter for balancing \(\mathcal{L}_\mathrm{cls}\) and \(\mathcal{R}_\mathrm{int}\), and \(\ell_\mathrm{CE}\) is cross-entropy loss.
Through this objective, the text decoder \(g_\phi\) is trained to focus on the textual features that are useful for minimizing \(\mathcal{L}_\mathrm{cls}\) while keeping the interpretability by \(\mathcal{R}_\mathrm{int}\).
We found that \(g_\phi\) easily collapses their output without \(\mathcal{R}_\mathrm{int}\).
Thus, the design of \(\mathcal{R}_\mathrm{int}\) is crucial for training XBMs.
However, since we often do not have the ground truth explanation sets in a real-world target dataset \(\mathcal{D}\), we can not directly penalize \(g_\phi\) with supervised losses as \(\mathcal{R}_\mathrm{int}\).
To overcome this challenge, we introduce a distillation-based approach using pre-trained text decoders in the next section.\looseness-1

\subsection{Explanation Distillation}
XBMs utilize pre-trained multi-modal models as the initial parameters of the text (explanation) decoder \(g_\phi\).
As an auto-regressive sequence model, the pre-trained text decoder \(g_{\phi_\mathrm{p}}\) can learn a conditional distribution \(q(\bm{e}|x)\) as 
\begin{equation}
     q(\bm{e}|x) = \prod^{L}_{l=1}q(e_l|x,\bm{e}_{<l}),\label{eq:pre_dist}
\end{equation}
where \(L\) is the maximum token length, \(e_l\) is the \(l\)-th token, and \(\bm{e}_{<l}\) is the text sequence before \(e_l\).
Since \(g_{\phi_\mathrm{p}}\) is trained on large-scale text-image pairs, \(q(\bm{e}|x)\) is expected to be able to generate a token sequence describing important information of various inputs \(x\).\looseness-1

Our key idea is to leverage \(q(\bm{e}|x)\) as the reference distribution for maintaining the interpretability of the generated explanation \(\hat{\bm{e}}\sim p_\phi(\bm{e}|x)\), where \(p_\phi(\bm{e}|x)\) is the model distribution of \(g_\phi\).
If \(p_\phi(\bm{e}|x)\) and \(q(\bm{e}|x)\) are sufficiently close, it can be guaranteed that the interpretability of the sequence generated by \(p_\phi(\bm{e}|x)\) approximate to that by \(q(\bm{e}|x)\).
Concretely, we compute the KL divergence between \(p_\phi(\bm{e}|x)\) and \(q(\bm{e}|x)\) as the regularization term \(\mathcal{R}_\mathrm{int}\) in Eq.~(\ref{eq:obj_xbm}).
\begin{eqnarray}
     \mathcal{R}_\mathrm{int}(\phi, \psi) &= D_\mathrm{KL}(q\|p_\phi) = \sum_{\bm{e}\in\mathcal{E}} q(\bm{e}|x)\log\left(\frac{q(\bm{e}|x)}{p_\phi(\bm{e}|x)}\right)\nonumber\\
     &= \mathbb{E}_{\bm{e}\sim q(\bm{e}|x)} \log\left(\frac{q(\bm{e}|x)}{p_\phi(\bm{e}|x)}\right).\label{eq:regularization_ideal}
\end{eqnarray}
However, \(D_\mathrm{KL}(q\|p_\phi)\) is computationally intractable because it requires multiple sequential sampling over \(\mathcal{E}=\mathcal{V}^L\) from \(q(\bm{e}|x)\) and the back-propagation through all sampling processes of \(p_\phi(e_l|x,\bm{e}_{<l})\).
To approximate Eq.~(\ref{eq:regularization_ideal}), we focus on the connection to knowledge distillation~\cite{hinton_2015_distilling}.
That is, minimizing Eq.~(\ref{eq:regularization_ideal}) can be seen as a knowledge distillation from \(g_{\phi_\mathrm{p}}\) to \(g_\phi\).
In such a sense, the approximation is
\begin{equation}
     \mathcal{R}_\mathrm{int}(\phi, \psi) \approx -\sum_{\bm{e}\in\mathcal{E}}\mathbb{I}_{\bm{e}=\bm{e}_\mathrm{p}}\log p_\phi(\bm{e}|x) = -\log p_\phi(\bm{e}=\bm{e}_\mathrm{p}|x),\label{eq:regularization_approx}
\end{equation}
where \(\bm{e}_\mathrm{p}\) is the sample from \(q(\bm{e}|x)\) and \(\mathbb{I}\) is the indicator function returning one when \(\bm{e}\) equals to \(\bm{e}_\mathrm{p}\) or returning zero otherwise; we omit the constant terms from the approximation for the simplicity.
As a concrete procedure, we first generate \(\bm{e}_\mathrm{p}\) from \(g_{\phi_\mathrm{p}}\) and then penalize the output logits of \(g_\phi\) through the cross-entropy loss for each output token in a next token prediction task.
This approximation technique is well-known as sequence-level knowledge distillation~\cite{Kim_EMNLP16_seqkd} in the field of neural machine translation, and it works well in the knowledge distillation of auto-regressive sequence models.
Sequence-level knowledge distillation corresponds to matching the modes of \(p\) and \(q\) and omits to transfer the uncertainty represented by the entropy \(H(q)\)~\cite{Kim_EMNLP16_seqkd}.
Nevertheless, we consider that this is sufficient for XBMs because the goal of XBMs is to provide interpretable explanations for target task predictions, not to replicate the pre-trained models perfectly.
We call the regularization with Eq.~(\ref{eq:regularization_approx}) \textit{explanation distillation}, and introduce it in training XBMs to maintain the text generation capability.

    \begin{algorithm}[t]
        \caption{Training of XBMs}\label{alg:training}
        \begin{algorithmic}[1]
        {\footnotesize
            \REQUIRE{Training dataset \(\mathcal{D}\), vision encoder \(h_\psi\), text decoder \(g_\phi\), classifier \(f_\theta\), pre-trained parameters \((\phi_\mathrm{p}, \theta_\mathrm{p})\), training batchsize \(B\), step size \(\eta\), trade-off parameter \(\lambda\)
            \ENSURE{Trained models \((h_\psi, g_\phi, f_\theta)\)}}
            \STATE{\texttt{\color{gray} \# Initialize parameters}}
            \STATE{\(\phi\leftarrow\phi_\mathrm{p},~~\psi\leftarrow\psi_\mathrm{p}\)}
            \WHILE{not converged}
            \STATE{\(\{(x^i,y^i)\}^B_{i=1}\sim \mathcal{D}\)}
            \STATE{\texttt{\color{gray}\# Generating reference explanation}}
            \STATE{\(\{\bm{e}^i_\mathrm{p}\}^B_i \leftarrow \{\operatorname{generate}(g_{\phi_\mathrm{p}}, h_\mathrm{p}(x^i))\}^B_i\)}
            \STATE{\texttt{\color{gray} \# Gumbel-softmax sampling}}
            \STATE{\(\{\hat{\bm{e}}^i\}^B_i \leftarrow \{\operatorname{g\_sampling}(g_\phi, h_\psi(x^i))\}^B_i\)}
            \STATE{\texttt{\color{gray}\# Computing batch-mean losses}}
            \STATE{\(\mathcal{L}^B_\mathrm{cls} \leftarrow \frac{1}{B}\sum^B_{i=1}\ell_\mathrm{CE}(f_\theta(h_\psi(x^i), \hat{\bm{e}}^i), y^i)\)}
            \STATE{\(\mathcal{R}^B_\mathrm{int} \leftarrow \frac{1}{B}\sum^B_{i=1}\ell_\mathrm{CE}(g_\phi\circ h_\psi(x^i), \bm{e}^i_\mathrm{p})\)}
            \STATE{\texttt{\color{gray}\# Updating parameters via backprop.}}
            \STATE{\(\theta \leftarrow \theta - \eta\nabla_\theta(\mathcal{L}^{B}_\mathrm{cls} + \lambda\mathcal{R}^{B}_\mathrm{int})\), \(\phi \leftarrow \phi - \eta\nabla_\phi(\mathcal{L}^{B}_\mathrm{cls} + \lambda\mathcal{R}^{B}_\mathrm{int})\), \(\psi \leftarrow \theta - \eta\nabla_\psi(\mathcal{L}^{B}_\mathrm{cls} + \lambda\mathcal{R}^{B}_\mathrm{int})\)}
            \ENDWHILE
        }
        \end{algorithmic}
    \end{algorithm}

\subsection{Algorithm}\label{sec:algorithm}
\paragraph{Training}
We show the training procedure in Algorithm~\ref{alg:training}.
In the training loop, we first generate the reference and predicted explanations \(\bm{e}_\mathrm{p}\) and \(\hat{\bm{e}}\) by \(\operatorname{generate}(\cdot)\) and \(\operatorname{g\_sampling}(\cdot)\), respectively (line 4 and 5).
To approximate the mode of \(q(\bm{e}|x)\) and ensure the quality as the reference, we generate \(\bm{e}_\mathrm{p}\) from frozen \(g_{\phi_\mathrm{p}}\) by beam search following the previous work~\cite{Kim_EMNLP16_seqkd}.
For sampling \(\hat{\bm{e}}\), we introduce the Gumbel-softmax trick~\cite{Jang_ICLR17_gumbel_softmax} to retain the computation graph for the end-to-end training with back-propagation.
The \(l\)-th token can be approximately sampled by
\begin{equation}\label{eq:g_sampling}
    e_l = \operatorname{softmax}((\log (g_\phi(h_\psi(x)))+\mathbf{g})/\tau),
\end{equation}
where \(\mathbf{g}=\{\mathrm{g}_1, ..., \mathrm{g}_{|\mathcal{V}|}\}\) is a vector of length \(|\mathcal{V}|\) where each element is sampled from \(\operatorname{Gumbel}(0,1)\) and \(\tau\) is the temperature parameter.
Intuitively, the temperature \(\tau\) controls the diversity of the token outputs from \(g_\phi\); larger \(\tau\) stimulates more diverse outputs.
To obtain diverse and accurate tokens for describing input, we apply exponential annealing to the temperature values according to the training steps, i.e., \(\tau^{(i+1)}=\tau^{(0)}\exp{(-r_\mathrm{a}i)}\), where \(i\) and \(r_\mathrm{a}\) are training step and annealing rate.
This allows XBMs to focus on the diversity of the output tokens in the early training steps and on the quality in the later steps.
We evaluate this design choice in Appendix E.1.
After sampling \(\bm{e}_\mathrm{p}\) and \(\hat{\bm{e}}\), we update all trainable parameters according to the objective function Eq.~(\ref{eq:obj_xbm}).

\begin{figure*}[t]
    \centering
    \includegraphics[width=\linewidth]{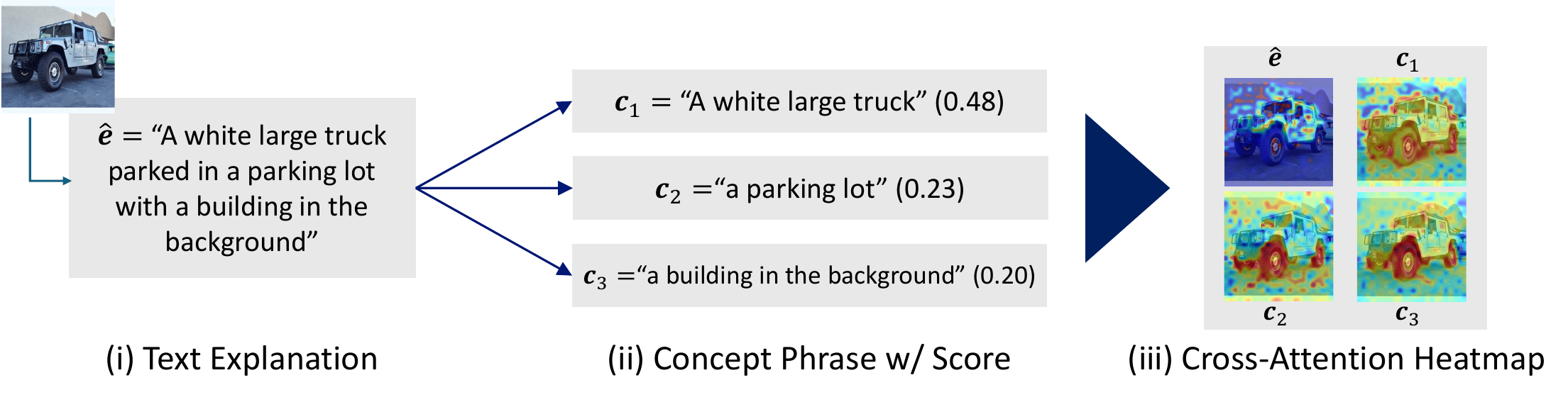}
    \vspace{-2.5mm}
    \caption{
    Explanation styles provided by XBMs. XBMs can output (i) text explanation directly generated from the explanation decoder, (ii) concept phrases with self-attention scores in the classifier, and (iii) cross-attention heatmap for the entire text explanation and each concept phrase.
    Concept phrases are constructed by a natural language parser, and the self-attention scores are computed in a middle layer of the classifier with respect to the \texttt{[CLS]} token for each concept phrase.
    Cross-attention heatmaps are the heatmap visualizations of cross-attention scores between input text tokens and image embedding tokens in the middle layer of the multi-modal classifier (a redder means a higher score).
    }
    \label{fig:explanation_styles}
    \vspace{-2.5mm}
\end{figure*}

\paragraph{Inference}
For the inference of test input \(x\), we generate \(\hat{\bm{e}}\) by beam search instead of the Gumbel-softmax trick, i.e., \(\hat{\bm{e}} \leftarrow \operatorname{generate}(g_\phi, h_\psi(x))\).
Finally, we return the target label prediction \(\hat{y}\leftarrow f_\theta(h_\psi(x), \hat{\bm{e}})\) and the explanation \(\hat{\bm{e}}\) to users.
Optionally, XBMs provide the other styles of explanation in addition to \(\hat{\bm{e}}\) (Fig.~\ref{fig:explanation_styles}). 
A \textit{concept phrase} \(\bm{c}\) is a noun phrase that compose \(\hat{\bm{e}}\), which can be extracted by natural language parser automatically~\cite{Feng_ICLR23_training-free_compositional_t2i}.
Similar to the concept outputs of CBMs, \(\bm{c}\) provides contributions of noun phrases in text explanations for the prediction.
For example, if the classifier \(f_\theta\) is implemented with transformer families with attention layers, we can interpret the contribution of \(\bm{c}\) for the target prediction \(\hat{y}\) via its self-attention scores as in Fig.~\ref{fig:explanation_styles} (ii).
Furthermore, we can visualize the cross-attention scores between the text explanations and visual tokens as a heatmap, suggesting what the model perceives as a concept in input data (Fig.~\ref{fig:explanation_styles} (iii)).

\section{Experiment}
We evaluate XBMs on multiple visual classification tasks and pre-training models.
We conduct qualitative and quantitative experiments on the explanation outputs of XBMs to evaluate the target performance and the interpretability.
We also provide a more detailed analysis, including varying hyperparameters \(\lambda, \tau\) and comparing explanation distillation with an alternative regularization loss in Appendix~E.

\subsection{Setting}\label{sec:exp_setting}
\paragraph{Implementation}
Our basic implementation of XBMs is based on BLIP~\cite{Li_ICML22_BLIP} because of its simplicity; we denote this model as XBM-BLIP.
That is, as the visual encoder \(h_\psi\), we used the ViT-B/32~\cite{Dosovitskiy_ICLR21_ViT}.
For the classifier \(f_\theta\), we used a BERT-base transformer~\cite{Devlin_NAACL19_BERT}; we input $h_\psi(x)$ into the cross-attention layers when using a multi-modal classifier inspired by BLIP~\cite{Li_ICML22_BLIP}.
We initialized \(\phi\) and \(\psi\) by the BLIP model pre-trained on image captioning tasks in the official repository\footnote{{\texttt{model\_base\_caption\_capfilt\_large.pth}} in \url{https://github.com/salesforce/BLIP}}.
We also report the results using larger pre-trained multi-modal models of LLaVA~\cite{liu_NeurIPS23_llava}.
We used v1.5 and v1.6 of LLaVA with multiple language model backbones (LLaMA2-7B~\cite{Touvron_arxiv23_llama2},  Vicuna-7B~\cite{vicuna2023}, and Mistral-7B~\cite{Jiang_arXiv23_mistral}); we denote these models as XBM-LLaVA.
We provide detailed training settings in Appendix~A.

\paragraph{Baselines}
We compare XBMs to black-box and interpretable baselines in performance and interpretability.
\textbf{Fine-tuned BLIP-ViT} is the black-box baseline, which directly optimizes the visual encoder of BLIP via fine-tuning.
\textbf{Label-free CBM}~\cite{Oikarinen_ICLR23_label-free_CBMs} is a state-of-the-art concept bottleneck model, which automatically constructs pre-defined concept sets from ConceptNet~\cite{Speer_AAAI17_conceptnet} or GPT-3~\cite{brown_NIPS20_gpt3} and then constructs concept embedding matrix via CLIP vision and text encoder.
We used BLIP-ViT as the backbone vision encoder of label-free CBMs.
\textbf{Frozen BLIP} baselines use frozen BLIP to generate text explanations and predict final labels by a multi-modal \(f_\theta(h_\psi(x), \hat{\bm{e}})\) or text classifier \(f_\theta(\hat{\bm{e}})\).
We also show the results of \textbf{XBM w/o \(\mathcal{R}_\mathrm{int}\)}, which updates \(g_\phi\) only on the classification loss Eq.~(\ref{eq:cls_loss}).

\begin{table*}[t]
    \centering
    \caption{Performance and Interpretability Evaluation of XBMs on multiple target datasets.}
    \label{tb:ex_multiple_target_blip}
    \vspace{-2.5mm}
        \begin{minipage}[t]{\textwidth}
        \resizebox{\textwidth}{!}{
        \begin{tabular}{lcccccc}\toprule
             & \multicolumn{3}{c}{Aircraft} & \multicolumn{3}{c}{Bird} \\\cmidrule(lr){2-4}\cmidrule(lr){5-7}
             & Test Acc. (\(\uparrow\)) & CLIP-Score (\(\uparrow\)) & GPT-2 Perplexity (\(\downarrow\)) & Test Acc. (\(\uparrow\)) & CLIP-Score (\(\uparrow\)) & GPT-2 Perplexity (\(\downarrow\)) \\
            \midrule
            Fine-tuned BLIP-ViT & 77.86$\pm$.30 & N/A & N/A & 83.48$\pm$.15 & N/A & N/A \\\midrule
            Label-free CBM (ConceptNet) & 15.37$\pm$.17 & 0.5356 & N/A & 17.67$\pm$.40 & 0.6025 & N/A \\
            Label-free CBM (GPT-3) & 44.47$\pm$.34 & 0.6153 & N/A & 77.74$\pm$.43 & 0.6904 & N/A \\
            \midrule
            Frozen BLIP +  $f_\theta(h_\psi(x), \hat{\bm{e}})$ & 45.23$\pm$.32 & 0.6824 & 155.8 & 68.03$\pm$.10 & 0.7535 & 173.5 \\
            XBM w/o \(\mathcal{R}_\text{int}\) & 70.78$\pm$.48 & 0.4730 & 322.6 & 61.94$\pm$.13 & 0.5137 & 431.0 \\
            XBM (Ours) & \textbf{74.09$\pm$.07} & \textbf{0.7151} & \textbf{129.8} & \textbf{80.99$\pm$.18} & \textbf{0.7942} & \textbf{166.8} \\
            \bottomrule
        \end{tabular}
        }
        \end{minipage}
        \begin{minipage}[t]{\textwidth}
            \resizebox{\textwidth}{!}{
        \begin{tabular}{lcccccc}\toprule
             & \multicolumn{3}{c}{Car} & \multicolumn{3}{c}{ImageNet}\\\cmidrule(lr){2-4}\cmidrule(lr){5-7}
             & Test Acc. (\(\uparrow\)) & CLIP-Score (\(\uparrow\)) & GPT-2 Perplexity (\(\downarrow\)) & Test Acc. (\(\uparrow\)) & CLIP-Score (\(\uparrow\)) & GPT-2 Perplexity (\(\downarrow\)) \\
            \midrule
            Fine-tuned BLIP-ViT & 90.08$\pm$.35 & N/A & N/A & 65.21$\pm$.14 & N/A & N/A\\\midrule
            Label-free CBM (ConceptNet) & 15.27$\pm$.13 & 0.5561 & N/A & 60.07$\pm$.42 & 0.6826 & N/A\\
            Label-free CBM (GPT-3) & 77.91$\pm$.21 & 0.6091 & N/A & 64.28$\pm$.09 & 0.7026 & N/A \\
            \midrule
            Frozen BLIP +  $f_\theta(h_\psi(x), \hat{\bm{e}})$  & 80.53$\pm$.29 & 0.6555 & 168.8 & 56.04$\pm$.49 & 0.7732 & 199.5 \\
            XBM w/o \(\mathcal{R}_\text{int}\) & 86.59$\pm$.11 & 0.4792 & 415.3 & 66.58$\pm$.30 & 0.5020 & 517.1 \\
            XBM (Ours)  & \textbf{89.47$\pm$.10} & \textbf{0.7173} & \textbf{131.8} & \textbf{67.83$\pm$.33} & \textbf{0.7920} & \textbf{122.8}\\
            \bottomrule
        \end{tabular}
        }
        \end{minipage}
        \vspace{-2.5mm}
\end{table*}

\paragraph{Datasets}
We used four image datasets for classification tasks in various domains: \textbf{Aircraft}~\cite{maji_13_aircraft}, \textbf{Bird}~\cite{Welinder_10_cub2002011}, \textbf{Car}~\cite{krause_3DRR2013_stanford_cars}, and \textbf{ImageNet}~\cite{russakovsky_imagenet}.
Aircraft, Bird, and Car are fine-grained image datasets, and ImageNet is a large-scale general image dataset.
For datasets other than ImageNet, we randomly split a dataset into \(9:1\) and used the former as the training set and the latter as the validation set.
For ImageNet, we set the split ratio \(99:1\) and used the official validation set as the test dataset.

\paragraph{Evaluation Metrics}
We report test accuracy as the target task performance.
For the interpretability evaluations, we introduce \textbf{CLIP-Score}~\cite{Radford_ICML21_CLIP,Hessel_EMNLP21_clipscore}, which is based on the cosine similarity between image embeddings and text embeddings on CLIP, i.e., higher is better.
CLIP-score was originally used to evaluate image captioning based on the relevance of the output captions to the input images.
Since it is highly sensitive to the hallucinations in the captions as reported in~\cite{Hessel_EMNLP21_clipscore}, CLIP-score can be used to assess the factuality of explanations.
For XBMs, we measured averaged CLIP-Scores between test inputs and the output explanations.
For Label-free CBMs, we measured averaged CLIP-Scores between test inputs and the output concept texts with the binary output of the concept bottleneck layer greater than 0.05; this threshold follows~\citet{Oikarinen_ICLR23_label-free_CBMs}.
We also introduce \textbf{GPT-2 Perplexity} as a measure of fluency in XBM's output explanations.
In general, perplexity scores on language models are calculated by the averaged cross-entropy of the next token probabilities and thus represent the fluency of the generated texts because the lower perplexity means that the sentence is composed of words that are likely to occur probabilistically.
Inspired by \citet{Chan_EMNLP23_clair}, we computed perplexity scores of explanations on GPT-2~\cite{Radford_Pre19_GPT2}.
That is, the generated explanations are unbiasedly evaluated by an external language model.
GPT-2 perplexity is helpful as a metric of the fluency of explanations because it shows the proximity to the natural text distribution learned by GPT-2.
We used open-sourced GPT-2 in huggingface transformers~\cite{Wolf_arXiv19_huggingface} to maintain reproducibility.\looseness-1

\begin{table*}[t]
    \centering
        \caption{
            Qualitative evaluation of explanation outputs.
        }\label{tb:qualitative}
        \vspace{-2.5mm}
        \resizebox{\linewidth}{!}{
        \begin{tabular}{lcc}\toprule
            & Bird & ImageNet \\
            & (Yellow Bellied Flycatcher) & (Lynx) \\
          &\begin{minipage}[t]{0.2\textwidth}
                \centering
                \includegraphics[align=c,width=1.0\columnwidth]{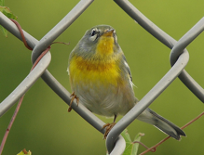}
          \end{minipage} &
          \begin{minipage}[t]{0.2\textwidth}
                \centering
                \includegraphics[align=c,width=1.0\columnwidth]{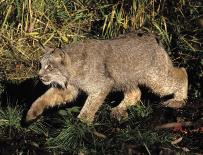}
          \end{minipage}\\\midrule
          Pre-trained BLIP (Caption) & 
          \begin{tabular}{c}
               {\small A bird perched on a wire fence with leaves}\\
               {\small on the ground and a blurry background.}
          \end{tabular} &
          \begin{tabular}{c}
               {\small Cat walking through the grass in}\\
               {\small the woods at night with it's eyes open.}
          \end{tabular}
          \\\midrule
          Label-free CBMs (Top-3 Concept) & 
          \begin{tabular}{c}
               {\small olive-colored sides (0.77)}\\
               {\small green head (0.55)}\\ 
               {\small a small, green body (0.52)}
          \end{tabular} &
          \begin{tabular}{c}
               {\small feline (0.98)}\\
               {\small long, sharp claws (0.53)}\\ 
               {\small mau (0.17)}
          \end{tabular}
          \\\midrule
        XBMs w/o $\mathcal{R}_\mathrm{int}$ (Text Explanation) &
          \begin{tabular}{c}
               {\small 2222222222222}\\
               {\small 2222222222}
          \end{tabular} &
          \begin{tabular}{c}
               {\small when when when when when }\\
               {\small when when when when when }
          \end{tabular}\\\midrule
          XBMs (Text Explanation) &
          \begin{tabular}{c}
               {\small A small green and yellow bird perched on }\\
               {\small a wire fence with leaves on the side.}
          \end{tabular} &
          \begin{tabular}{c}
               {\small Furry feline walking in the woods at night }\\
               {\small with its eyes open and one paw on the ground.}
          \end{tabular}\\\midrule
          XBMs (Top-3 Concept Phrase) & 
          \begin{tabular}{c}
               {\small a small green and yellow bird (0.39)}\\
               {\small leaves on the side (0.32)}\\
               {\small a wire fence (0.21)}
          \end{tabular} &
          \begin{tabular}{c}
               {\small Furry feline (0.39)}\\
               {\small one paw on the ground (0.19)}\\ 
               {\small the woods (0.17)}
          \end{tabular}
          \\\midrule
          XBMs (Cross-Attn. Heatmap) & 
          \begin{minipage}[t]{0.2\textwidth}
                \centering
                \includegraphics[align=c,width=1.0\columnwidth]{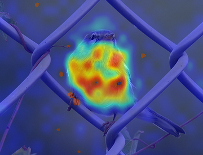}
          \end{minipage} &
          \begin{minipage}[t]{0.2\textwidth}
                \centering
                \includegraphics[align=c,width=1.0\columnwidth]{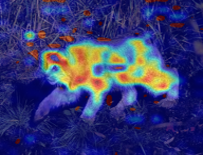}
          \end{minipage}
          \\\bottomrule
        \end{tabular}
        }
 \end{table*}

\subsection{Design Evaluation of XBMs}\label{sec:exp_ablation_xbm}

\subsubsection{Quantitative Evaluation}\label{sec:exp_quantitative}
Table~\ref{tb:ex_multiple_target_blip} demonstrates the quantitative performance and interpretability of XBM-BLIP on the four target datasets.
For the target performance, our XBMs outperformed the Label-free CBM baselines and achieved competitive performance with the black-box baseline in the test accuracy.
In particular, XBM achieved high performance on datasets where label-free CBM did not perform well (i.e., Aircraft and Car).
This can be caused by insufficient pre-defined concepts due to the limited vocabulary in ConceptNet and GPT-3 about describing objects in these datasets, whereas XBMs promote multi-modal understanding by training the explanation decoder to describe arbitrary objects useful for the target dataset with unlimited vocabulary.
For the interpretability, XBMs outperformed CBMs in CLIP-Score.
This indicates that the explanations from XBMs are more factual to the input images than the concept outputs of CBMs, which are in pre-defined concept sets.

Furthermore, the ablation study in the bottom rows of Table~\ref{tb:ex_multiple_target_blip} shows that the objective function in Eq.~(\ref{eq:obj_xbm}) works effectively as we expected.
Compared to the frozen BLIP baselines, which simply apply fixed pre-trained BLIP to generate text captions, our XBM significantly improved all of the test accuracy, CLIP-Score, and GPT-2 Perplexity.
This suggests that optimizing text decoders with respect to target tasks guides the generated explanation to be informative and target-related for solving the task.
We also confirm that the regularization term \(\mathcal{R}_\mathrm{int}\) by explanation distillation (Eq.~(\ref{eq:regularization_approx})) is crucial to generate meaningful explanation; XBM w/o \(\mathcal{R}_\mathrm{int}\) catastrophically degraded CLIP-Score and GPT-2 Perplexity.

\subsubsection{Qualitative Evaluation}\label{sec:exp_qualitative}
Table~\ref{tb:qualitative} shows the qualitative studies of explanations generated from XBMs; we also show the other examples in Appendix~B.
We computed the self-/cross attention scores in the middle of the transformer layers by following \citet{Zhang_ICLR20_BERTScore}.
For comparison, we also show the top-3 concept outputs of CBMs and the generated captions of pre-trained BLIP, i.e., the initial states of XBMs.
The text explanations of XBMs contain more detailed information than pre-trained BLIP.
This is because the target classification loss \(\mathcal{L}_\mathrm{cls}\) forces the text decoders to describe target-related visual information to solve the task.
Importantly, XBMs without explanation distillation \(\mathcal{R}_\mathrm{int}\) generate totally broken explanations, indicating the objective function of XBMs succeed in training the models to focus on the tokens related to the target task without the collapse of explanations.
Meanwhile, the concept phrase explanations show the contributions to the final outputs (i.e., self-attention scores) for each noun phrase in the text explanations.
In contrast to CBM's concepts, the concept phrases tend to be aligned with visual features appearing in input images rather than describing input by pre-defined knowledge.
This is easy for humans to understand when interpreting the output of the models.
Finally, the cross-attention heatmaps intuitively localize where the generated text explanations correspond to the input image spaces.
We confirm that the heatmaps concentrate on objects through optimization and facilitate a multi-modal understanding of the image in Section~\ref{sec:exp_analysis_cross_attn_heatmap}.

\begin{figure*}[t]
    \centering
    \includegraphics[width=\linewidth]{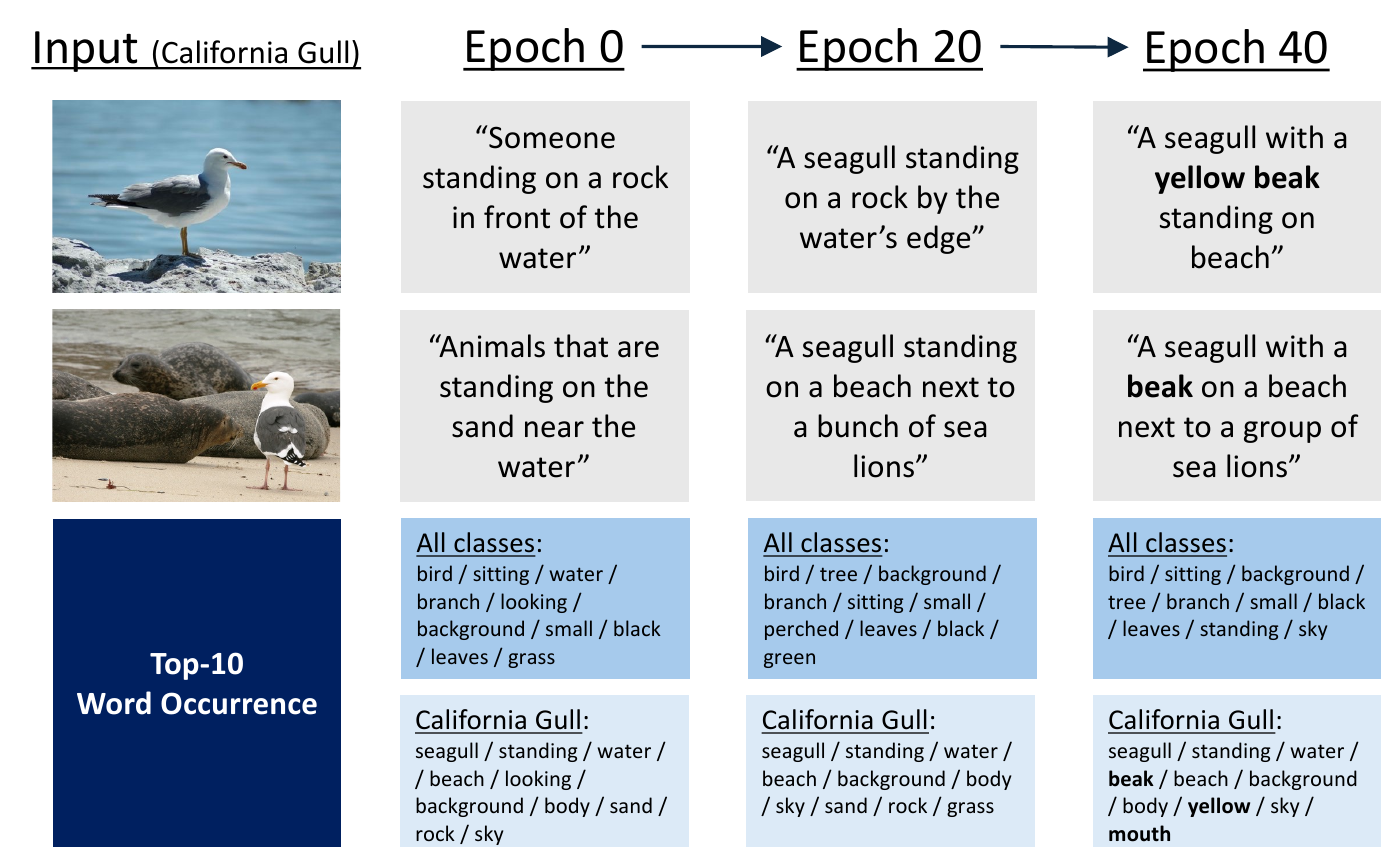}
    \vspace{-2.5mm}
    \caption{
    Transition of XBM's explanation outputs during training (please zoom in).
    }
    \label{fig:explanation_transition}
\end{figure*}

We also analyze the transition of the generated explanations in Fig.~\ref{fig:explanation_transition}.
We print the text explanation of XBMs and the top-10 word occurrence for all classes and the input class at 0, 20, and 40 epochs.
According to the training epoch, the explanations and words progressively focus on detailed and target-related information in images.
Concretely, in this example, the XBM is optimized to describe ``yellow beak (mouth)'', a key feature of California Gull.
These suggest that XBMs can provide interpretable and useful explanations for humans.\looseness-1

\subsection{XBMs with Large Vision-Language Models}\label{sec:exp_large_models}
Here, we evaluate the scalability and practicality of XBMs by combining them with larger vision-language models than BLIP.
Instead of BLIP, we used the LLaVA models with various language model backbones~\cite{liu_NeurIPS23_llava}.
Table~\ref{tb:ex_llm} shows that leveraging the high-performance vision-language model in XBMs yields better performance and interpretability scores, suggesting that the XBM's objective function can enhance the multi-modal understanding ability even if using the large vision-language models pre-trained on massive image-text pairs.
This emphasizes the flexibility of XBM, consisting of arbitrary vision-language models.

\subsection{XBMs with Text Classifier}\label{sec:exp_text_cls}
Table~\ref{tb:ex_llm} also evaluates XBMs with a text classifier \(f_\theta(\hat{\bm{e}})\), which relies only on text information for the final predictions.
Although XBM-BLIP with \(f_\theta(\hat{\bm{e}})\) drops the performance from one with a multi-modal classifier \(f_\theta(h_\psi(x), \hat{\bm{e}})\), switching the backbone from BLIP to LLaVA~\cite{liu_NeurIPS23_llava} resolves the performance gap.
This indicates that more sophisticated vision-language models make XBMs generate informative text explanations, and they can achieve practical performance even when not using input features \(h_\psi(x)\).
Appendix~C further shows the results on the other datasets.

\subsection{Evaluations of Cross-Attention Heatmap}\label{sec:exp_analysis_cross_attn_heatmap}
The cross-attention heatmap explanation of XBMs visualizes the local input space regions correlated to the text explanation in the classifier.
To assess the validity of XBMs on improving multi-modal understanding, we evaluate the generated heatmaps on the ImageNet segmentation task by following \citet{Chefer_CVPR21_transformer_interpretability} and \citet{Gandelsman_ICLR24_interpreting_CLIP}.
That is, we generate the heatmaps on the test set of ImageNet Segmentation~\cite{Guillaumin_IJCV14_imagenet_segmentation} and compute the pixel accuracy, mean IoU (mIoU), and mean average precision (mAP) with the ground truth segmentation masks.
Through this evaluation, we can evaluate how heatmaps cover the object of target classes in the pixel spaces.
Table~\ref{tb:ex_imagenetseg} shows the results.
Compared to the frozen BLIP, XBM-BLIP improved all of the segmentation metrics.
This means that the training objective of XBMs encourages the multi-modal understanding of target class objects on the models.
In Appendix~D, we further compare the XBM's heat maps with existing attribution methods, such as GradCAM~\cite{Selvaraju_ICCV17_gradcam}.

\begin{table*}[t]
    \centering
    \caption{Evaluation of XBMs with text and multi-modal classifiers built on large vision-language models on ImageNet.} 
    \vspace{-2.5mm}
    \label{tb:ex_llm}
        \resizebox{\linewidth}{!}{
        \begin{tabular}{lcccccc}\toprule
            & \multicolumn{3}{c}{Text Classifier \(f_\theta(\bm{e})\)} & \multicolumn{3}{c}{Multi-modal Classifier \(f_\theta(h_\psi(x), \bm{e})\)}\\\cmidrule(lr){2-4}\cmidrule(lr){5-7}
            & Test Acc. (\(\uparrow\)) & CLIP-Score (\(\uparrow\)) & GPT-2 Perplexity (\(\downarrow\)) & Test Acc. (\(\uparrow\)) & CLIP-Score (\(\uparrow\)) & GPT-2 Perplexity (\(\downarrow\)) \\\midrule
            Frozen BLIP                & 9.97$\pm$.12  & 0.7732 & 199.5 & 56.04$\pm$.49 & 0.7732 & 199.5 \\
            XBM-BLIP                   & 18.26$\pm$.31 & 0.8007 & 148.1 & 67.83$\pm$.33 & 0.7920 & 122.8 \\
            Frozen LLaVA-v1.5-LLaMA-7B & 64.01$\pm$.46 & 0.7773 & 236.8 & 70.21$\pm$.18 & 0.7773 & 100.8\\
            XBM-LLaVA-v1.5-LLaMA-7B    & 71.41$\pm$.25 & 0.8008 & 127.2 & 72.95$\pm$.16 & 0.7998 & 82.6\\
            XBM-LLaVA-v1.6-Vicuna-7B   & 73.73$\pm$.30   & 0.8140    & 36.74   & 74.42$\pm$.23 & 0.8037 & 32.3\\
            XBM-LLaVA-v1.6-Mistral-7B  & 72.14$\pm$.27   & 0.8037    & 20.67   & 74.04$\pm$.11 & 0.8130 & 21.7\\
            \bottomrule
        \end{tabular}
        }
\end{table*}

\begin{table}[t]
    \centering
    \centering
    \caption{Evaluation of cross-attention map of XBMs on ImageNet Segmentation.}
    \label{tb:ex_imagenetseg}
        \resizebox{\columnwidth}{!}{
        \begin{tabular}{lccc}\toprule
            & Pixel Acc. (\(\uparrow\)) & mIoU (\(\uparrow\)) & mAP (\(\uparrow\))\\\midrule
            Frozen BLIP + Multi-modal Classifier & 78.67 & 57.90 & 79.72 \\
            XBM-BLIP & 80.90 & 60.80 & 80.18\\
            \bottomrule
        \end{tabular}
        }
\end{table}

\subsection{Reliability Evaluation via Human Intervention}\label{sec:exp_intervention}
CBMs allow the debugging of the model behavior through human intervention in the predicted concepts~\cite{Koh_ICML20_concept_bottleneck}.
Similarly, we can debug the behavior of XBMs by intervening in the generated explanations.
Here, we show examples of an intervention in which all explanations are replaced to check the effect of the explanation quality on the final classification results.
At inference, we replace the generated explanations from the explanation decoder with modified explanations.
We tested two types of interventions: (i) randomized and (ii) ground-truth explanations.
For randomized explanation, we used a token sequence uniformly sampled from the vocabulary space for the length of the originally generated explanation.
For ground-truth explanation, we used the extended annotation set for Bird proposed by~\citet{Reed_CVPR16_visual_desc}.
Table~\ref{tb:ex_intervention} shows the performance of the intervened XBM-BLIP models.
The intervened explanations with randomized explanations significantly degraded the performance of XBM-BLIP, indicating that the generated explanations are essential to achieving high performance.
In contrast, the intervention with ground-truth explanations largely improved the performance. This suggests that higher-quality explanations can yield higher performance, and intervening with human explanations is helpful for XBMs to improve their performance.
In other words, the final prediction of XBMs largely depends on the content of the generated explanation \(\hat{\bm{e}}\), indicating that \(\hat{\bm{e}}\) is a reliable explanation for the final prediction.
To conclude, these results support the debuggability of XBMs and the reliability of the generated explanations.

\section{Related Work}\label{sec:relatedwork}
The main research directions of the interpretability of black-box deep neural networks are briefly divided into attribution-based and concept-based methods.
Attribution-based methods such as CAM~\cite{Zhou_CVPR16_cam} and GradCAM~\cite{Selvaraju_ICCV17_gradcam} generate a localization map representing important regions for the model predictions for specific classes.
However, since the maps generated by attribution-based methods do not have information other than that they responded to the predictions, they are less interpretable regarding what semantic input features contribute to the output.
In contrast to these methods, our XBMs can generate semantically interpretable heatmaps via cross-attention between image and text explanations, which can be decomposed at the level of noun phrases.

On the other hand, concept-based methods such as TCAV~\cite{Kim_ICML18_tcav} and CBMs~\cite{Koh_ICML20_concept_bottleneck} compute contribution scores for pre-defined concepts on intermediate outputs of models.
Among them, CBMs are highly relevant to our XBMs since both have interpretable intermediate layers in models.
CBMs predict concept labels and then predict final class labels from the predicted concepts.
The original CBMs have the challenge of requiring human annotations of concept labels~\cite{Zarlenga_NeurIPS22_concept_embedding,Moayeri_ICML23_text-to-concept,Xu_ICLR24_energy-based_CBMs}.
Post-hoc CBMs~\cite{Yuksekgonul_ICLR23_post-hoc_CBMs} and Label-free CBMs~\cite{Oikarinen_ICLR23_label-free_CBMs} addressed this challenge by automatically collecting concepts corresponding to target task labels by querying large language models (e.g., GPT-3~\cite{Brown_NeurIPS20_GPT3}) or existing concept banks (e.g., ConceptNet~\cite{Speer_AAAI17_conceptnet}).
However, CBMs' explanations are still restricted to pre-defined concepts, and they are not necessarily reliable because CBMs often predict the concepts without mapping to corresponding input regions~\cite{Huang_AAAI24_concept_trustworthiness}.
On the contrary, our XBMs directly generate natural language explanations to interpret the model outputs without pre-defined concepts.

Similar to our work, a few works attempted to generate linguistic explanations for target classification models~\cite{Hendricks_ECCV16_generating_explanation,Nishida_ACL22_improving_by_description}.
However, these methods require ground truth text explanations for training models, which are expensive and restrict applications.
Our XBMs address this limitation by learning explanation generation by the classification loss and explanation distillation using a pre-trained text decoder.

\section{Limitation}\label{sec:limitation}
One of the limitations of XBMs is that they can not generate explanations based on user-defined concepts, which can be expressed by CBMs.
In other words, XBMs are good at fluently explaining outputs in a general vocabulary because of their language model backbone but have difficulty giving interpretations for fixed concepts based on expert knowledge.
A promising direction of future work is to associate the fluent explanations with user-defined concepts.

\section{Conclusion}
In this paper, we presented a novel interpretable deep neural networks called explanation bottleneck models (XBMs).
By leveraging pre-trained vision-language models, XBMs generate explanations corresponding to input and output in the forms of natural language description, concept phrases with contribution scores, and cross-attention heatmaps on input spaces.
To ensure both the target task performance and the explanation quality, XBMs are optimized by the target task loss with explanation distillation, which penalizes the divergence between the distributions of the training and pre-trained text decoders.
Experiments show that XBMs can achieve both high target task performance and accurate and fluent explanations; they achieve competitive performance to black-box baselines and largely outperform CBMs in target test accuracy.
Furthermore, we found that training of XBMs can enhance the multi-modal understanding capability of backbone vision-language models even when using large vision-language models pre-trained on massive image-text pairs.
We believe that this work introduces a new perspective on natural language explanations and advances the study of interpretable deep models to the next paradigm.

\begin{table}[t]
    \centering
    \centering
    \caption{Evaluation of Intervened XBMs on Bird.}
    \label{tb:ex_intervention}
        \resizebox{\columnwidth}{!}{
        \begin{tabular}{lccc}\toprule
            & Test Acc. (\(\uparrow\)) & CLIP-Score (\(\uparrow\)) & GPT-2 Perplexity (\(\downarrow\))\\\midrule
            XBM-BLIP & 80.99 & 0.7942 & 166.8\\
            Intervened XBM-BLIP (Randomized) & 44.42 & 0.4497 & 4631.1\\
            Intervened XBM-BLIP (Ground-Truth) & 82.21 & 0.8179 & 104.5\\
            \bottomrule
        \end{tabular}
        }
\end{table}

\section*{Acknowledgement}
We thank the members of the Kashima Laboratory at Kyoto University and our NTT colleagues, especially Han Bao and Daiki Chijiwa, for their helpful feedback on this research.

\clearpage

\appendix
\section*{Appendix}
\section{Detailed Setting}\label{sec:append_detailed_setting}
\subsection{Training}\label{sec:append_training_detail}
We trained the models by the AdamW~\cite{Loshchilov_ICLR19_AdamW} optimizer with the initial learning rate of 3.0\(\times\)10\(^{-5}\) that decayed by cosine annealing.
The training epochs were 100 on the Aircraft/Bird/Car datasets and 5 on the ImageNet dataset.
We used mini-batch sizes of 32.
The input samples were resized into resolutions of \(384\times384\) for XBM-BLIP and \(336\times336\) for XBM-LLaVA according to the setting of vision encoders.
We used \(\lambda\) of \(0.1\) and \(\tau\) of 10 with exponential annealing by \(r_\mathrm{a}=1.0\times10^{-4}\) if not otherwise noted; we discuss the effect of \(\lambda\) and \(\tau\) in Section~\ref{sec:exp_analysis}.
For the experiments on XBM-LLaVA, we fine-tuned the LoRA adapter parameters~\cite{Hu_ICLR22_lora} of backbone language models instead of the entire parameters.
We selected the final model by checking the validation accuracy for each epoch.
We implemented the training and evaluation with PyTorch-1.13.
We ran the experiments three times on a 24-core Intel Xeon CPU with eight NVIDIA A100 GPUs with 80GB VRAM and recorded the average evaluated on the final models; we omit the standard deviations for saving spaces, but we have confirmed the statistical significance of our method with a p-value \(<0.05\) toward baselines.

\subsection{Datasets}\label{sec:append_dataset_detail}
{\bf ImageNet} \citep{russakovsky_imagenet}: We downloaded ImageNet from the official site \url{https://www.image-net.org/}.
ImageNet is released under a license that allows it to be used for non-commercial research/educational purposes (see \url{https://image-net.org/download.php}).

{\bf Aircraft (FGVC Aircraft)} \citep{maji_13_aircraft}: We downloaded FGVC Aircraft from the official site \url{https://www.robots.ox.ac.uk/~vgg/data/fgvc-aircraft/}.
FGVC Aircraft is released under a license that allows it to be used for non-commercial research/educational purposes (see \url{https://www.robots.ox.ac.uk/~vgg/data/fgvc-aircraft/}).

{\bf Bird (CUB-200-2011)} \citep{Welinder_10_cub2002011}: We downloaded CUB-200-2011 from the official site \url{http://www.vision.caltech.edu/datasets/cub_200_2011/}.
CUB-200-2011 is released under a license that allows it to be used for non-commercial purposes (see \url{https://authors.library.caltech.edu/27452/}).

{\bf Car (Stanford Cars)} \citep{krause_3DRR2013_stanford_cars}: We downloaded Stanford Cars from the official site\url{https://ai.stanford.edu/~jkrause/cars/car_dataset.html}.
StanfordCars is released under a license that allows it to be used for non-commercial research purposes (see \url{https://ai.stanford.edu/~jkrause/cars/car_dataset.html}).

\section{Additional Qualitative Experiments}\label{sec:append_qualitative_experiments}
Table~\ref{tb:appendix_qualitative} shows the qualitative evaluation results on the Aircraft and Car datasets, which are omitted in the main paper due to the page constraint.
The evaluation protocol is the same as Section~\ref{sec:exp_qualitative}.
Similar to Table~\ref{tb:qualitative}, our method succeeded in capturing the semantic concepts of input images in the text explanation.
Also, the concept phrases and cross-attention heatmaps show that the captured semantic concepts contribute to the final output and the main focus of models is on the target objects.

\begin{table*}[t]
    \centering
        \caption{
            Qualitative evaluation of explanation outputs.
        }\label{tb:appendix_qualitative}
        \resizebox{\textwidth}{!}{
        \begin{tabular}{lcc}\toprule
            & Aircraft & Car\\
            & (ATR-42) & (Hummer)\\
          &\begin{minipage}[t]{0.2\textwidth}
                \includegraphics[align=c,width=\columnwidth]{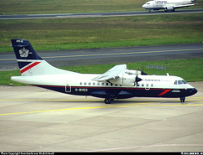}
          \end{minipage}&
          \begin{minipage}[t]{0.2\textwidth}
                \includegraphics[align=c,width=\columnwidth]{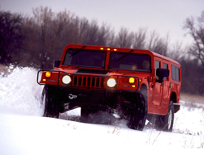}
          \end{minipage}
          \\\midrule
          Pre-trained BLIP (Caption) & 
          \begin{tabular}{c}
               {\small These two planes are parked}\\
               {\small on the tarmac at an airport run way.}
          \end{tabular} &
          \begin{tabular}{c}
               {\small Someone is driving a red jeep on a snowy}\\
               {\small road with trees in the background.}
          \end{tabular}
          \\\midrule
          Label-free CBMs (Top-3 Concept) & 
          \begin{tabular}{c}
               {\small canard foreplanes (0.70)}\\
               {\small a single-engine propeller (0.64)}\\ 
               {\small fixed landing gear (0.58)}
          \end{tabular} &
        \begin{tabular}{c}
               {\small 2500HD model designation (0.63)}\\
               {\small off-road tires (0.40)}\\ 
               {\small distinct Suzuki grille (0.39)}
          \end{tabular}
          \\\midrule
          XBMs (Text Explanation) &
          \begin{tabular}{c}
               {\small A small white and blue airplane on a runway at}\\
               {\small an airport with another plane in the background.}
          \end{tabular} &
          \begin{tabular}{c}
               {\small A truck with off-road tires driving through the snow}\\
               {\small in the wintertime with trees in the background.}
          \end{tabular}
          \\\midrule
          XBMs (Top-3 Concept Phrase) & 
          \begin{tabular}{c}
               {\small another plane in the background (0.34)}\\
               {\small a small white and blue airplane (0.31)}\\
               {\small a runway at an airport (0.22)}
          \end{tabular} &
        \begin{tabular}{c}
               {\small a truck with off-road tires (0.36)}\\
               {\small trees in the background (0.26)}\\ 
               {\small the wintertime (0.23)}
          \end{tabular}
          \\\midrule
          XBMs (Cross-Attn. Heatmap) & 
          \begin{minipage}[t]{0.2\textwidth}
                \includegraphics[align=c,width=\columnwidth]{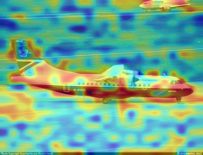}
          \end{minipage}&
          \begin{minipage}[t]{0.2\textwidth}
                \includegraphics[align=c,width=\columnwidth]{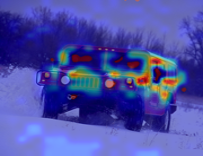}
          \end{minipage}
          \\\bottomrule
        \end{tabular}
        }
 \end{table*}

\section{Additional Results with Large Vision-Language Models}~\label{sec:append_llava}
In Sections~\ref{sec:exp_large_models}~and~\ref{sec:exp_text_cls}, we confirm the results of XBMs combined with LLaVA  (XBM-LLaVA) on ImageNet.
Here, we show the additional XBM-LLaVA results on the other datasets.
Table~\ref{tb:append_ex_llm} demonstrates the results with the same tendency as Table~\ref{tb:ex_llm}, i.e., combining XBMs with larger vision-language backbones significantly improves the target task performance and interoperability.
This indicates that our methods can be extendable even if a new and powerful vision-language model emerges.

\begin{table*}[t]
    \centering
    \caption{Evaluation of XBMs with large vision-language models}
    \label{tb:append_ex_llm}
        \resizebox{\linewidth}{!}{
        \begin{tabular}{lcccccc}\toprule
            \textbf{\textit{Aircraft}} & \multicolumn{3}{c}{Text Classifier \(f_\theta(\bm{e})\)} & \multicolumn{3}{c}{Multi-modal Classifier \(f_\theta(h_\psi(x), \bm{e})\)}\\\cmidrule(lr){2-4}\cmidrule(lr){5-7}
             & Test Acc. (\(\uparrow\)) & CLIP-Score (\(\uparrow\)) & GPT-2 Perplexity (\(\downarrow\)) & Test Acc. (\(\uparrow\)) & CLIP-Score (\(\uparrow\)) & GPT-2 Perplexity (\(\downarrow\)) \\\midrule
            Frozen BLIP                & 3.95  & 0.6824 & 155.8 & 45.23 & 0.6824 & 155.8 \\
            XBM-BLIP                   & 24.36 & 0.7084 & 145.9 & 74.09 & 0.7151 & 129.8 \\
            Frozen LLaVA-v1.5-LLaMA-7B & 59.21 & 0.7500 & 227.4 & 73.03 & 0.7514 & 227.3\\
            XBM-LLaVA-v1.5-LLaMA-7B    & 64.11 & 0.7515 & 179.5 & 78.77 & 0.7595 & 184.8 \\
            XBM-LLaVA-v1.6-Vicuna-7B   & 68.64 & 0.7842 & 22.9 & 82.08 & 0.7758 & 34.7 \\
            XBM-LLaVA-v1.6-Mistral-7B  & 67.06 & 0.7769 & 39.9 & 81.55 & 0.7851 & 21.9 \\
            \bottomrule
        \end{tabular}
        }
        \resizebox{\linewidth}{!}{
        \begin{tabular}{lcccccc}\toprule
            \textbf{\textit{Bird}} & \multicolumn{3}{c}{Text Classifier \(f_\theta(\bm{e})\)} & \multicolumn{3}{c}{Multi-modal Classifier \(f_\theta(h_\psi(x), \bm{e})\)}\\\cmidrule(lr){2-4}\cmidrule(lr){5-7}
             & Test Acc. (\(\uparrow\)) & CLIP-Score (\(\uparrow\)) & GPT-2 Perplexity (\(\downarrow\)) & Test Acc. (\(\uparrow\)) & CLIP-Score (\(\uparrow\)) & GPT-2 Perplexity (\(\downarrow\)) \\\midrule
            Frozen BLIP                & 5.53  & 0.7535 & 173.5 & 68.03 & 0.7535 & 173.5 \\
            XBM-BLIP                   & 19.52 & 0.7910 & 168.5 & 80.99 & 0.7942 & 166.9 \\
            Frozen LLaVA-v1.5-LLaMA-7B & 75.20 & 0.7788 & 140.8 & 75.67 & 0.7788 & 107.3 \\
            XBM-LLaVA-v1.5-LLaMA-7B    & 80.87 & 0.7981 & 107.3 & 83.07 & 0.8037 & 21.8  \\
            XBM-LLaVA-v1.6-Vicuna-7B   & 82.33 & 0.8130 & 21.0  & 84.93 & 0.8154 & 21.6 \\
            XBM-LLaVA-v1.6-Mistral-7B  & 81.53 & 0.8101 & 16.7  & 84.73 & 0.8110 & 16.7 \\
            \bottomrule
        \end{tabular}
        }
        \resizebox{\linewidth}{!}{
        \begin{tabular}{lcccccc}\toprule
            \textbf{\textit{Car}} & \multicolumn{3}{c}{Text Classifier \(f_\theta(\bm{e})\)} & \multicolumn{3}{c}{Multi-modal Classifier \(f_\theta(h_\psi(x), \bm{e})\)}\\\cmidrule(lr){2-4}\cmidrule(lr){5-7}
             & Test Acc. (\(\uparrow\)) & CLIP-Score (\(\uparrow\)) & GPT-2 Perplexity (\(\downarrow\)) & Test Acc. (\(\uparrow\)) & CLIP-Score (\(\uparrow\)) & GPT-2 Perplexity (\(\downarrow\)) \\\midrule
            Frozen BLIP                & 7.98  & 0.6091 & 168.8 & 77.91 & 0.6091 & 168.8 \\
            XBM-BLIP                   & 27.57 & 0.7127 & 168.5 & 89.47 & 0.7173 & 131.8 \\
            Frozen LLaVA-v1.5-LLaMA-7B & 83.50 & 0.7236 & 99.8  & 91.19 & 0.7300 & 97.2 \\
            XBM-LLaVA-v1.5-LLaMA-7B    & 86.18 & 0.7300 & 97.2  & 92.82 & 0.7322 & 83.8  \\
            XBM-LLaVA-v1.6-Vicuna-7B   & 86.70 & 0.8081 & 35.7  & 93.85 & 0.8032 & 41.4 \\
            XBM-LLaVA-v1.6-Mistral-7B  & 86.75 & 0.8086 & 25.8  & 92.41 & 0.8071 & 27.9 \\                                                                                                                                                                                                                                                                                                          
            \bottomrule
        \end{tabular}
        }
    \vspace{-2.5mm}
\end{table*}

\section{Additional Results of ImageNet Segmentation}~\label{sec:append_imagenetseg}
We additionally compare our method with existing attribution methods on BLIP-ViT. 
Note that we omit this result from the main paper because, strictly speaking, BLIP-ViT and XBM-BLIP are different models and thus this evaluation is not a direct comparison.
By following Chefer et al.~\cite{Chefer_CVPR21_transformer_interpretability}, we tried LRP~\cite{Binder_ICANN16_lrp}, partial-LRP~\cite{Voita_ACL19_partial_lrp}, rollout~\cite{Abnar_ACL20_rollout}, raw attention output from BLIP-ViT, GradCAM~\cite{Selvaraju_ICCV17_gradcam}, and the method of~\cite{Chefer_CVPR21_transformer_interpretability}.
Table~\ref{tb:append_imagenetseg} shows the results of ImageNet Segmentation with the same setting of Table~\ref{tb:ex_imagenetseg}.
Surprisingly, the cross-attention output of the classifier \(f_\theta\) (i.e., Pre-trained BLIP and XBM-BLIP) significantly outperformed the conventional visualization methods in the segmentation metric.
This indicates that visualization explanation outputs of XBMs are quite accurate and reliable as the interpretation of model outputs.

\begin{table*}[t]
    \centering
    \centering
    \caption{Evaluation of cross-attention map of XBMs on ImageNet Segmentation.}
    \label{tb:append_imagenetseg}
        \begin{tabular}{lccc}\toprule
            & Pixel Acc. (\(\uparrow\)) & mIoU (\(\uparrow\)) & mAP (\(\uparrow\))\\\midrule
            LRP~\cite{Binder_ICANN16_lrp}(BLIP-ViT) & 46.25 & 29.69 & 48.51 \\
            partial-LRP~\cite{Voita_ACL19_partial_lrp} (BLIP-ViT) & 53.59 & 36.29 & 65.06 \\
            rollout~\cite{Abnar_ACL20_rollout} (BLIP-ViT) & 52.73 & 35.81 & 66.78 \\
            Raw Attention (BLIP-ViT) & 57.12 & 39.00 & 67.92 \\
            GradCAM~\cite{Selvaraju_ICCV17_gradcam} (BLIP-ViT) & 61.84 & 39.68 & 63.48 \\
            Chefer et al.~\cite{Chefer_CVPR21_transformer_interpretability} (BLIP-ViT) & 59.92 & 42.30 & 69.51 \\\midrule
            XBM-BLIP w/ Fixed Decoder & 78.67 & 57.90 & 79.72 \\
            XBM-BLIP & 80.90 & 60.80 & 80.18\\
            \bottomrule
        \end{tabular}
\end{table*}

\section{Detailed Analysis}\label{sec:exp_analysis}
In this section, we provide detailed analyses of XBMs.
In particular, we assess temperature annealing in the Gumbel softmax sampling (Eq.~(\ref{eq:g_sampling})), the hyperparameter \(\lambda\) in Eq.~(\ref{eq:obj_xbm}), and the localization ability of the cross-attention heatmaps introduced in Section~\ref{sec:algorithm}.

\subsection{Evaluations of Temperature Annealing}\label{sec:exp_analysis_tannealing}
We introduce the temperature annealing strategy for determining \(\tau\) in Eq.~(\ref{eq:g_sampling}).
Here, we evaluate the effects by varying the initial temperature \(\tau^{(0)}\) in \(\{1,10,100\}\).
Table~\ref{tb:analysis_temp} shows the test performance and interpretability scores.
We tested the cases leveraging a constant temperature \(\tau^{(0)}\) and applying exponential temperature annealing, i.e., \(+\) Annealing.
In the cases of constant temperatures, we confirm that the larger temperatures tend to achieve better target performance but degrade perplexity scores.
This is because using a larger temperature increases the entropy of the generative distribution of tokens in the Gumbel softmax sampling, and thus, it slightly loses the naturalness of the generated sentences.
On the other hand, applying the temperature annealing improved all scores in all initial temperatures.
This implies that, by gradually reducing the temperature, XBMs can try to generate diverse tokens in the early stages of learning, and it narrows down only vocabulary with high likelihood in the later stages while the sentence naturalness is maintained.

\subsection{Effects of Hyperparameter \(\lambda\)}\label{sec:exp_analysis_hyperparams}
The hyperparameter \(\lambda\) in Eq~(\ref{eq:obj_xbm}) balances the target task training and the regularization to avoid the collapse of text explanations. 
Table~\ref{tb:analysis_lambda} shows the results when varying \(\lambda\).
It demonstrates that the cases of \(\lambda>0\) can avoid the collapse of the interpretability and improve target performance.
We see that there is a trade-off between the target test accuracy and the GPT-2 Perplexity scores.
In contrast, fortunately, CLIP-Score was less sensitive to the value of \(\lambda>0\), suggesting high-performance XBMs can still generate explanations well-related to inputs.
Therefore, we recommend determining \(\lambda\) based on whether fluency or accuracy is a priority according to the application's requirements.

\begin{table*}[t]
    \centering
    \caption{Effects of temperature \(\tau\) and annealing for Gumbel softmax sampling of XBMs (Car).}
    \label{tb:analysis_temp}
        \begin{tabular}{lccc}\toprule
            & Test Acc. (\(\uparrow\)) & CLIP-Score (\(\uparrow\)) & GPT-2 Perplexity (\(\downarrow\))\\\midrule
            \(\tau^{(0)}=1\) & 86.60 & 0.7138 & 140.8\\
            \quad\(+\) Annealing & 87.12 & 0.7258 & 127.3 \\
            \(\tau^{(0)}=10\) & 86.71 & 0.7148 & 143.9 \\
            \quad\(+\) Annealing & 88.65 & 0.7253 & 133.0\\
            \(\tau^{(0)}=100\) & 88.03 & 0.7168 & 146.2\\
            \quad\(+\) Annealing & 88.56 & 0.7272 & 143.6\\
            \bottomrule
        \end{tabular}
\end{table*}

\begin{table*}[t]
    \centering
    \caption{Effects of hyperparameter \(\lambda\) of XBMs (Car).}
    \label{tb:analysis_lambda}
        \begin{tabular}{lccc}\toprule
            & Test Acc. (\(\uparrow\)) & CLIP-Score (\(\uparrow\)) & GPT-2 Perplexity (\(\downarrow\))\\\midrule
            \(\lambda=0\)  & 86.59 & 0.4792 & 415.3\\\midrule
            \(\lambda=0.01\) & 89.18 & 0.7158 & 163.4\\
            \(\lambda=0.1\)  & 89.47 & 0.7172 & 145.4\\
            \(\lambda=0.3\)  & 89.09 & 0.7148 & 138.9\\
            \(\lambda=0.5\)  & 88.62 & 0.7167 & 132.6\\
            \(\lambda=0.7\)  & 87.58 & 0.7158 & 131.7\\
            \(\lambda=1.0\)  & 87.59 & 0.7138 & 127.3\\
            \bottomrule
        \end{tabular}
    \vspace{-2.5mm}
\end{table*}

\subsection{Explanation Distillation vs. Other Regularization}\label{sec:exp_analysis_explanation_distill}
Here, we evaluate our explanation distillation regularization \(\mathcal{R}_\mathrm{int}\) through a comparison to another regularization method.
We compare our explanation distillation to L2SP~\cite{xuhong_ICML18_l2sp}, which penalizes the model parameters by minimizing the l2 distance from the pre-trained parameters.
Table~\ref{tb:analysis_exp_distill} shows the results on the Car dataset.
We confirm that our explanation distillation outperforms L2SP in all performance metrics. 
Our method largely improves clip score, while L2SP degrades it from frozen BLIP.
These results suggest that directly regularizing the decoder's output helps XBMs explore the vocabulary needed for a task through classification loss while preserving natural sentences and minimizing the gap in the parameter space, which is harmful to this purpose.

\begin{table*}[t]
    \centering
    \caption{Effects of Regularization in XBMs (Car).}
    \label{tb:analysis_exp_distill}
        \begin{tabular}{lccc}\toprule
            & Test Acc. (\(\uparrow\)) & CLIP-Score (\(\uparrow\)) & GPT-2 Perplexity (\(\downarrow\))\\\midrule
            Frozen BLIP  & 77.91 & 0.6091 & 168.8 \\
            Explanation Distillation (Ours)  & 90.48 & 0.7173 & 131.8 \\
            L2SP~\cite{xuhong_ICML18_l2sp} & 87.47 & 0.5059 & 159.4 \\
            \bottomrule
        \end{tabular}
    \vspace{-2.5mm}
\end{table*}

\section{Broader Impacts}\label{sec:append_broader_impacts}
A potential negative effect introduced by our work is that XBMs may output biased explanations if the backbone language model is extremely biased.
This can be avoided by purifying the language model with existing debiasing methods such as~\cite{Kaneko_EACL21_debiasing} before training XBMs.
Since the target tasks handled by XBMs are no different from those in general models, off-the-shelf defence methods may be directly applicable to other risks such as adversarial attacks.

\newpage
\bibliography{ref}

\end{document}